\documentclass{article} 
\usepackage{nfam2025_workshop}
\usepackage{times}
\usepackage{enumitem}

\usepackage{amsmath,amsfonts,bm}

\def\eqref#1{equation~\ref{#1}}

\def\1{\bm{1}}

\DeclareMathAlphabet{\mathsfit}{\encodingdefault}{\sfdefault}{m}{sl}
\SetMathAlphabet{\mathsfit}{bold}{\encodingdefault}{\sfdefault}{bx}{n}

\usepackage{hyperref}
\usepackage{url}
\usepackage{graphicx}

\title{Word2Spike: Poisson Rate Coding for Associative Memories and Neuromorphic Algorithms}

\author{Archit Kalra* \\
Department of Bioengineering\\
Rice University\\
Houston, TX 77030 USA \\
\texttt{\{archit.kalra\}@rice.edu} \\
\And
Midhun Sadanand \\
Department of Computer Science \\
Yale University \\
New Haven, CT 06511 USA \\
\texttt{\{midhun.sadanand\}@yale.edu} \\
}

\iclrfinalcopy

\begin{document}

\maketitle

\begin{abstract}
Spiking neural networks offer a promising path toward energy-efficient, brain-like associative memory. This paper introduces Word2Spike, a novel rate coding mechanism that combines continuous word embeddings and neuromorphic architectures. We develop a one-to-one mapping that converts multi-dimensional word vectors into spike-based attractor states using Poisson processes. Using BitNet b1.58 quantization, we maintain 97\% semantic similarity of continuous embeddings on SimLex-999 while achieving 100\% reconstruction accuracy on 10,000 words from OpenAI’s text-embedding-3-large. We preserve analogy performance (100\% of original embedding performance) even under intentionally introduced noise, indicating a resilient mechanism for semantic encoding in neuromorphic systems. Next steps include integrating the mapping with spiking transformers and liquid state machines (resembling  Hopfield Networks) for further evaluation.
\end{abstract}

\section{Introduction}

Associative memory networks aim to link features into high-dimensional representations that can reconstruct context from partial information \citep{KrotovHopfield2016, Krotov2023, hoover2023energy}. While biological neural systems are good at performing such tasks, artificial neural networks, especially “reasoning models” like GPT o1, struggle with episodic memory, even with short tasks (10k-100k tokens) \citep{huet2025episodic}. Spiking neural networks (SNNs) offer a promising alternative, more closely mimicking biological information processing through temporal and rate-based coding mechanisms, as well as having utility for recurrent associative memory models \citep{Maass1997, ravichandran2024spiking}. However, effective input coding remains a challenge \citep{guo2021neural, zou2023toward, schuman2019non}.

Although prior work has demonstrated embedding binarization and conversion to neural spikes via temporal coding, where the time a spike occurs reflects information about the embedding, there is still risk of “embedding collapse” (where multiple words have the same embedding) without careful manipulation of the loss function to ensure each dimension represents a different domain \citep{tissier2018near, zhang2021faster}.We attempted to replicate past work by designing autoencoders, using sigma-delta quantization and performing Matryoshka Representation Learning to train binary embeddings from scratch, but found that embedding collapse was quite common (85\% uniqueness post-quantization), even with the use of uniqueness loss functions \citep{tissier2018near}.

Moreover, though temporal coding can relay considerable information, it is also prone to distortion in noisy environments \citep{park2021noise}. Rate coding, based on neuron firing rate, is also a biologically plausible learning rule \citep{gautrais1998rate}. Though temporal coding can achieve low latency, rate coding offers key advantages for associative memory formation: (1) Greater robustness to noise distortion and adversarial attacks, critical for stable attractor dynamics, (2) Better stability in deep architectures like Hopfield-inspired spiking transformers, and (3) Natural incorporation of stochasticity that mirrors biological memory systems \citep{guo2021neural, kim2022rate, ravichandran2024spiking, mueller2021spiking}. This stochasticity provides implicit regularization and helps avoid spurious attractors during learning \citep{ma2023exploiting}. Prior rate-coding approaches, however, have often relied on deterministic neurons, thereby neglecting the inherent Poisson–like stochasticity and noise in biological systems \citep{jiang2024stochastic}. 

In response to these issues, we believed there was a need for a stochastic rate-coding based embedding conversion model to allow for SNN training with word embeddings. We describe the implementation below, and the code can be accessed at \url{https://github.com/aka133/NeuromorphicML} for reproducibility.

\section{Method}
We design a one-to-one mapping scheme that allows for near-lossless embedding conversion between quantized vectors and spikes. The process is similar to the pulse method described by \citet{schuman2019non}, but with an additional quantization step. First, the value of each dimension of a given word vector is quantized using an absmean quantization technique as used in Microsoft’s BitNet b1.58, which maintained impressive accuracy while performing quantization down to -1, 0 and +1 for each continuous value \citep{ma2024era}:

Let $\vec{W} = [w_1, w_2, \ldots, w_n]$ denote a continuous word embedding. We find the mean absolute value:
\[
\gamma = \frac{1}{n}\sum_{i=1}^{n} |w_i|.
\]
Each dimension is then quantized as
\[
\tilde{W}_i = \begin{cases}
+1 & \text{if } w_i > \gamma, \\
0 & \text{if } |w_i| \leq \gamma, \\
-1 & \text{if } w_i < -\gamma.
\end{cases}
\]
After quantization, we assign one spiking neuron per dimension and define its firing rate $\nu_i$ by
\[
\nu_i = \begin{cases}
100\,\text{Hz} & \text{if } \tilde{W}_i = +1, \\
0\,\text{Hz}   & \text{if } \tilde{W}_i = 0, \\
50\,\text{Hz}  & \text{if } \tilde{W}_i = -1.
\end{cases}
\]

Neural spikes are modeled as stochastic events drawn from a Poisson distribution with mean $\nu_i$. Spikes are generated over a fixed observation window (e.g., 200\,ms), and the reverse mapping counts spikes to estimate the firing rate. This leads to some noise (the rate in a given window is not always exactly 50 or 100 Hz), so we define a threshold for decoding (see Appendix 1 for justification):
\[
\hat{W}_i = \begin{cases}
+1 & \text{if } 72\,\text{Hz} \leq \nu_i \leq 100\,\text{Hz}, \\
0  & \text{if } \nu_i = 0\,\text{Hz}, \\
-1 & \text{if } 0\,\text{Hz} \leq \nu_i < 72\,\text{Hz}.
\end{cases}
\]
Note firing rates are set here with a 2:1 ratio, but this is arbitrary. This design yields nearly lossless reconstruction of the quantized embedding even in the presence of biologically realistic noise.

\begin{figure}[h]
\centering
\includegraphics[width=1.0\textwidth]{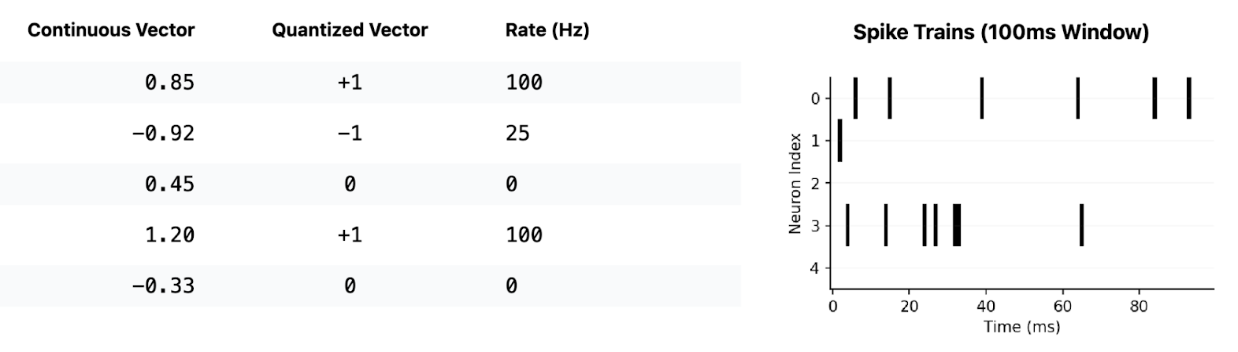}
\caption{A continuous word vector is normalized and quantized to $\{-1,0,+1\}$. The quantized values are mapped to firing rates (e.g., 25 Hz for $-1$, 0 Hz for 0, and 100 Hz for $+1$). Poisson–distributed spikes are generated, and the reverse process recovers the quantized vector.}
\end{figure}

\section{Evaluation}
We evaluated our embedding conversion pipeline using three different representations: Original continuous embeddings (obtained via OpenAI’s text-embedding-3-large); Quantized embeddings (obtained by applying an absmean-based discretization to each dimension); and Spike-based embeddings (obtained by converting the quantized embeddings into spike trains using a Poisson process over a 200 ms window, and then reconstructing the discrete vector using a tolerance of 22 Hz).

We collected the 10,000 most common words per Google’s 10,000 English dataset, then evaluated the models on four benchmarks: SimLex-999 word similarity, a custom analogy dataset to assess top-1 accuracy, nearest-neighbor consistency (evaluating overlap in top-10 nearest neighbors of each word between the embeddings) and reconstruction accuracy (ability of the spike raster to perfectly recover the quantized embedding).

\begin{table}[!htb]
\centering
\caption{Performance of Embedding Types on Evaluation Metrics}
\label{tab:performance}
\begin{tabular}{lccc}
\hline
\textbf{Metric} & \textbf{Original} & \textbf{Quantized} & \textbf{Spike–based} \\
\hline
SimLex–999 (Spearman’s $\rho$) & 0.540 & 0.542 & 0.526 \\
Analogy Accuracy (25 analogies) & 37.50\% & 37.50\% & 37.50\% \\
Nearest–Neighbor Consistency (Overlap@10) & 0.885 & 0.885 & 0.727 \\
Reconstruction Accuracy & N/A & N/A & 100.00\% \\
\hline
\end{tabular}
\end{table}

The quantized embeddings preserve semantic structure nearly perfectly, and despite the introduction of Poisson noise, the spike rasters retain sufficient semantic fidelity for downstream tasks.

\section{Discussion}
Our method achieves perfect reconstruction accuracy (100\%). Though the spike–based representation shows a modest drop ($\rho \approx 0.526$ from $\rho \approx 0.542$) due to Poisson noise, the relational structure required for analogical reasoning is maintained across all representations (analogy accuracy of 37.50\% for each). This consistency is key: it demonstrates that our pipeline maintains the same analogical reasoning capabilities as the original embeddings, even after quantization and conversion to spikes. This preservation indicates promise for neuromorphic associative memory systems.

The slight degradation in nearest–neighbor consistency is acceptable given that the introduced noise mimics biological spiking behavior. These findings support our claim that the proposed conversion pipeline preserves semantic content while enabling use of rate–coded spike representations in SNNs. 

Future work will compare our rate–coding technique with temporal coding schemes. Given the high semantic fidelity maintained by our method, we anticipate that it may outperform binary embeddings in neuromorphic applications. Additionally, we plan to study the effects of varying frequencies and time windows. Preliminary tests with 400\,ms windows (using 25\,Hz for $-1$ and 200\,Hz for $+1$) yielded promising results, including 100\% reconstruction accuracy, 98.5\% semantic similarity preservation, and 98.9\% nearest–neighbor consistency. 

Liquid State Machines (LSM), as introduced by \citet{MaassRealTimeComputing}, are particularly relevant for associative memory as they maintain a "memory trace" of past inputs through recurrent connections, functioning like a deep Hopfield network \citep{hasani2022liquid}. Since only the readout layer needs training, LSMs avoid backpropagation while still capturing temporal associations \citep{MaassRealTimeComputing}. Our initial experiments with rate-coded inputs for reservoir dynamics (see Appendix 2) suggest potential for multi-pattern associative recall, like hippocampal memory replay \citep{ponghiran2019reinforcement}.

\section{Conclusion}
We have introduced a novel 1:1 rate–coding mechanism that converts continuous word embeddings into spike–based representations through absmean quantization and Poisson spiking. Our evaluations confirm that—despite the introduction of realistic noise—the spike–based embeddings preserve the essential semantic and relational structure of the original embeddings. This work opens the door to future neuromorphic language models and EEG–integrated systems.

\newpage

\clearpage  

\bibliography{references}
\bibliographystyle{nfam2025_workshop}

\newpage

\appendix

\section{Appendix 1: Justification for Reverse Mapping Boundaries}
For a neuron firing at rate $\lambda$ Hz observed over a time window T, the number of spikes N follows a Poisson distribution with variance equal to $\lambda$T. Over our 200ms observation window, a 100 Hz neuron produces $N_{100} = 20 ± \sqrt{20}$ spikes while a 50 Hz neuron produces $N_{50} = 10 ± \sqrt{10}$ spikes. Converting back to rates, this yields variations of ±22.4 Hz at 100 Hz versus ±15.8 Hz at 50 Hz. The 72 Hz threshold places the decision boundary between these ranges while accounting for their different variances. The observation window length can also be adjusted based on application requirements, trading temporal resolution for statistical confidence.

\section{Appendix 2: Spiking Transformer and Liquid State Machine}
\label{app:spiking_models}

Spiking transformers (e.g., Spikformer, SpikeBERT, SpikeGPT) utilize linear spiking–self attention (SSA) for language modeling \citep{zhou2023spikformer, lv2023spikebert, zhu2023spikegpt}. Our preliminary experiments with a spiking transformer confirm that a forward pass using rate–coded embeddings is feasible. This transformer is untrained so the outputs are not yet coherent, but this proof–of–concept suggests that spiking transformers can generate language tokens using our conversion scheme. SNNs are universal approximators, so with sufficient training, we expect more coherent outputs \citep{zhang2022theoretically}.

\begin{figure}[h]
\centering
\fbox{
\begin{minipage}{0.8\textwidth}
\texttt{\textbf{Input text:} Artificial intelligence is transforming technology and society.}\\
\texttt{\textbf{Output text:} Lowering intellect marge marge marge lowering lowering marge}
\end{minipage}}
\caption{Random output from an undertrained spiking transformer. Although the outputs are not yet coherent, they can be decoded into words using surrogate gradients.}
\end{figure}

Liquid State Machines (LSM), as introduced by \citet{MaassRealTimeComputing}, function as reservoir computers where the internal spiking neurons need not be trained---only the output layer is. This approach is particularly appealing for SNNs, given the challenges of backpropagation through deep spiking networks \citep{Maass1997}. The code for both of these implementations can be accessed at \url{https://github.com/aka133/NeuromorphicML}.

\end{document}